\documentclass[conference]{IEEEtran}

\IEEEoverridecommandlockouts
\IEEEpubid{\makebox[\columnwidth]{Preprint; Accepted to iSAI-NLP 2025}
\hspace{\columnsep}\makebox[\columnwidth]{ }}
\usepackage{cite}
\usepackage{amsmath,amssymb,amsfonts}
\usepackage{algorithmic}
\usepackage{graphicx}
\usepackage{textcomp}
\usepackage{multirow}
\usepackage{url}
\usepackage{xcolor}
\usepackage{booktabs}
\usepackage[font=normalsize,labelfont=bf]{caption}

\usepackage{fontspec}
\newfontfamily\padauktext[Script=Myanmar]{Padauk-Regular.ttf}

\def\BibTeX{{\rm B\kern-.05em{\sc i\kern-.025em b}\kern-.08em
    T\kern-.1667em\lower.7ex\hbox{E}\kern-.125emX}}

\def\BibTeX{{\rm B\kern-.05em{\sc i\kern-.025em b}\kern-.08em
    T\kern-.1667em\lower.7ex\hbox{E}\kern-.125emX}}

\makeatletter
\newcommand{\linebreakand}{%
  \end{@IEEEauthorhalign}
  \hfill\mbox{}\par
  \mbox{}\hfill\begin{@IEEEauthorhalign}
}
\makeatother

\begin{document}

\title{ASR Error Correction in Low-Resource Burmese with Alignment-Enhanced Transformers using Phonetic Features}

    
     

\author{
Ye Bhone Lin$^{1,+}$,
Thura Aung$^{1,2,+}$,
Ye Kyaw Thu$^{1,3,*}$,
Thazin Myint Oo$^{1}$
\\
$^{1}$Language Understanding Laboratory, Myanmar
\\
$^{2}$Department of Computer Engineering, KMITL, Bangkok, Thailand  
\\
$^{3}$Language and Semantic Technology Research Team, NECTEC, Bangkok, Thailand
\\
\emph{Corresponding author:} \texttt{yekyaw.thu@nectec.or.th}$^{*}$
\\
\emph{Emails:} \texttt{yebhonelin10@gmail.com}, \texttt{66011606@kmitl.ac.th}, \texttt{queenofthazin@gmail.com}
\\
$^{*}$Corresponding author
\\
$^{+}$Equal contribution
}

\maketitle

\begin{abstract}
This paper investigates sequence-to-sequence Transformer models for automatic speech recognition (ASR) error correction in low-resource Burmese, focusing on different feature integration strategies including IPA and alignment information. To our knowledge, this is the first study addressing ASR error correction specifically for Burmese. We evaluate \textit{five} ASR backbones and show that our ASR Error Correction (AEC) approaches consistently improve word- and character-level accuracy over baseline outputs. 
The proposed AEC model, combining IPA and alignment features, reduced the average WER of ASR models from 51.56 to 39.82 before augmentation (and 51.56 to 43.59 after augmentation) and improving chrF++ scores from 0.5864 to 0.627, demonstrating consistent gains over the baseline ASR outputs without AEC.
Our results highlight the robustness of AEC and the importance of feature design for improving ASR outputs in low-resource settings.
\end{abstract}


\section{Introduction}
 
Automatic Speech Recognition (ASR) systems often generate transcription errors, which can negatively affect downstream applications such as machine translation and information retrieval \cite{kannan2018analysis}. To address this, ASR Error Correction (AEC) has been proposed as a post-processing step to improve transcription quality without modifying the acoustic model \cite{sodhi2021mondegreen}. Traditional AEC approaches relied on external language models for re-scoring ASR hypotheses \cite{tanaka2018neural}, while more recent approaches explore Large Language Models (LLMs) for generative error correction, showing advantages over conventional language models \cite{yang2023generative}. 

End-to-end AEC methods based on Sequence-to-Sequence (S2S) architectures have gained popularity as they directly map erroneous transcripts to ground-truth text \cite{mani2020asr, liao2023improving}. Beyond text-only methods, some studies incorporate both acoustic features and ASR hypotheses to enable cross-modal AEC \cite{mu2024mmgermultimodalmultigranularitygenerative, radhakrishnan2023whispering, chen2024never}. 

Recent advances highlight the effectiveness of S2S models; for example, \cite{dutta2022asr} use a pre-trained S2S BART model as a denoising system to correct phonetic and spelling errors, fine-tuned on both synthetic and real ASR errors, and rescored using word-level alignments, achieving significant WER improvements on accented speech.

Several ASR datasets and systems have been developed for Burmese, including UCSY-SC1 \cite{mon2019ucsy} and large-vocabulary systems covering data collection, lexicon construction, and acoustic and language modeling \cite{vocab_myanspeech}. MyanSpeech combines a CTC-Attention acoustic model with an RNN language model to improve alignment and recognition accuracy \cite{myan_speech}, while ChildASR targets primary-level students using a GMM-HMM model trained on ~5 hours of child speech \cite{myanchildasr}. Additionally, myMediCon introduces a Burmese medical speech corpus and evaluates Transformer and RNN-based ASR models \cite{mymedicon}. Despite having ASR datasets developed for Burmese, only two datasets are open-source corpora: OpenSLR80 \cite{oo-etal-2020-burmese} and FLEURS \cite{FLEURS2022arxiv}. In the previous studies, phonetic information has also been shown to improve recognition performance in tonal languages such as Burmese \cite{myanspeechtone}. 

Although existing systems improve acoustic modeling for low-resource Burmese, unlike post-OCR \cite{ocr} and spelling error correction \cite{spelling}, ASR outputs still contain errors that can significantly affect downstream applications, and post-ASR error correction remains largely unexplored. To address this gap, we propose an alignment-guided Transformer Sequence-to-Sequence model with phonetic features based on the International Phonetic Alphabet (IPA) for low-resource Burmese AEC. Since training data is limited, we apply speech-level data augmentation and generate synthetic ASR errors from the augmented audio, rather than augmenting post-ASR text directly. 

In this work, we present the AEC parallel corpus with data augmentation pipeline and a set of fine-tuned Whisper ASR models for Burmese general-domain speech recognition. These resources aim to facilitate research and development in low-resource Burmese ASR and automatic error correction.

\begin{figure*}[ht!]
    \centering
    \caption{(a) Post-ASR Dataset Preparation for ASR Error Correction (AEC) Training (b) Integration of Phonetic Features and Alignment in AEC.}
    \includegraphics[width=0.8\linewidth]{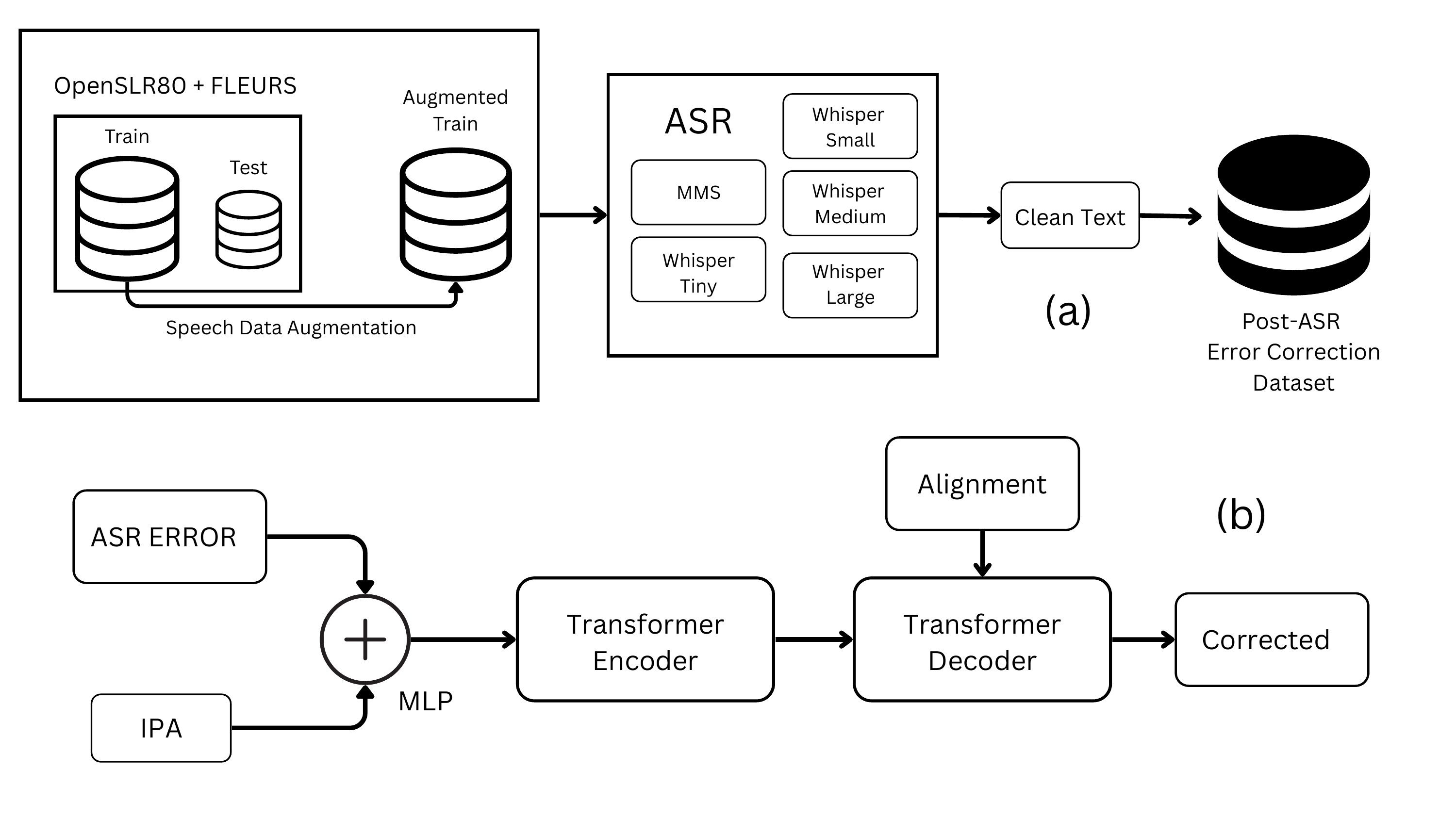} 
    \label{fig:framework}
\end{figure*}

\section{Dataset Preparation}
 
We used two open-source Burmese speech corpora: OpenSLR80\footnote{\url{chuuhtetnaing/myanmar-speech-dataset-openslr-80}} \cite{oo-etal-2020-burmese} and FLEURS\footnote{\url{chuuhtetnaing/myanmar-speech-dataset-google-fleurs}} \cite{FLEURS2022arxiv}. Table~\ref{tab:speech_data} summarizes the amount of training and test data in hours. We also report the MOSNet scores for each dataset, which provide an automatic estimate of perceived speech quality. Both datasets achieve good MOSNet scores (4.06–4.14), indicating that the recordings are suitable for ASR.

Table \ref{tab:speech_data} summarizes the amount of training and test data in hours. We also report the MOSNet\cite{mosnet} scores for each dataset, which provide an automatic estimate of perceived speech quality. To increase both the diversity and the amount of training data, we applied data augmentation to 10\% of the training set for each method. Waveform-based augmentations included pitch shifting, speed perturbation, loudness adjustment, background noise addition, temporal shifting, and random cropping. Spectrogram-based augmentations included Vocal Tract Length Perturbation (VTLP) and time/frequency masking. These techniques enrich acoustic variability and improve model robustness. We applied augmentation only to the training split, keeping the test split intact. The augmentation process was done using \texttt{NLPAug}\footnote{\url{https://github.com/makcedward/nlpaug}} Python library.

\begin{table}[h!]
\renewcommand{\arraystretch}{1.5}
    \centering
    \caption{Speech Data Information}

    \begin{tabular}{|c|r|r|r|}
        \hline
        \textbf{Dataset} & \textbf{Train Hr.} & \textbf{Test Hr.} & \textbf{MOSNet} \\
        \hline
        OpenSLR80 & 3.70 & 0.42 & 4.06 \\
        \hline
        FLEURS & 15.95 & 1.64 & 4.14 \\
        \hline
    \end{tabular}
     
    \label{tab:speech_data}
\end{table}
 
For ASR, we experimented with Massively Multilingual Speech (MMS) \cite{pratap2023mms} and Whisper \cite{radford2022whisper} models in different sizes (tiny, small, medium, and large). Pretrained Whisper models performed poorly on Burmese, so we fine-tuned them using the available speech data to improve transcription quality. To generate ASR errors for AEC, we used both pretrained MMS and fine-tuned Whisper, applied on original as well as augmented data. Both ground-truth transcripts and ASR outputs (with errors) were segmented into syllables using \texttt{myWord}\footnote{\url{https://github.com/ye-kyaw-thu/myWord}} \cite{thu2021myword}. Since ASR outputs often contained punctuation, unseen tokens, and special characters, we cleaned the text before feeding it into the error correction model. 

Figure \ref{fig:framework} (a) shows the process of dataset preparation for ASR Error Correction, and Table \ref{tab:dataset_syllables} shows the Post-ASR Dataset statistics: the number of sentences in parallel data and the total number of syllables (Syl) for original, augmented, and test sets, for both Err (ASR errors) and GT (ground truth).


\begin{table}[h!]
\centering
\renewcommand{\arraystretch}{1.5}
\setlength{\tabcolsep}{8pt}
\caption{Pretrained ASR model information: model, number of parameters, and language coverage. \textsuperscript{*} Whisper models were finetuned for Burmese in our work.}
\begin{tabular}{|l|c|r|}
\hline
\textbf{Model} & \textbf{Parameters} & \textbf{Languages} \\
\hline
MMS-1B ASR     & $\sim$1B   & 1,100+ \\
Whisper Tiny\textsuperscript{*}   & 39M        & 99  \\
Whisper Small\textsuperscript{*}  & 244M       & 99  \\
Whisper Medium\textsuperscript{*} & 769M       & 99  \\
Whisper Large\textsuperscript{*}  & 1.55B      & 99  \\
\hline
\end{tabular}
 
\label{tab:asr_models}
\end{table}

\begin{table}[h!]
\centering
\renewcommand{\arraystretch}{1.5}
\setlength{\tabcolsep}{10pt}
\caption{Post-ASR Dataset statistics: number of sentences in parallel data and total number of syllables (Syl) for original, augmented, and test sets for both Err (error) and GT (groundtruth).}
\begin{tabular}{|l|r|r|r|}
\hline
\textbf{Dataset} & \textbf{Sentences} & \textbf{Err Syl} & \textbf{GT Syl} \\
\hline
Original        & 31.7k   & 1.25M & 1.22M \\
Augmented       & 55.9k   & 2.17M & 2.19M \\
Test            & 3.19k   & 0.13M & 0.12M \\
\hline
\end{tabular}
 
\label{tab:dataset_syllables}
\end{table}

\vspace{3mm}
\section{Methodology}

\subsection{Feature Extraction}
\paragraph{IPA Features}
We incorporated Phonetic (IPA-based) features to improve the training of the correction model. To extract IPA, Grapheme-to-IPA (G2IPA) conversion, we trained sequence tagging models using the myG2P word-level dictionary version 2.0 \cite{htun2021grapheme}. Among the methods tested, Conditional Random Fields (CRF) outperformed both Ripple Down Rule (RDR) and BiLSTM models, as shown in Table \ref{tab:g2ipa_result}.

\begin{table}[h!]
 
\caption{Grapheme-to-IPA Conversion Results}
\renewcommand{\arraystretch}{1.5}
    \centering
    \begin{tabular}{|c|c|c|c|}
    \hline
        & \textbf{Precision} & \textbf{Recall} & \textbf{F1} \\
    \hline    
    RDR & 0.84 & 0.84 & 0.85 \\
    \hline
    CRF & \textbf{0.97} & \textbf{0.99} & \textbf{0.98} \\
    \hline
    BiLSTM & 0.93 & 0.92 & 0.93 \\
    \hline
    \end{tabular}
     
    \label{tab:g2ipa_result}
    
\end{table}
 
\paragraph{Alignment Feature} 
We used alignment to reduce hallucinations in the AEC models. For this purpose, we employed fast-align\footnote{\url{http://github.com/clab/fast_align}}
\cite{dyer-etal-2013-simple}, a log-linear re-parameterization of IBM Model 2 that addresses the strong assumptions of Model 1 and the over-parameterization of Model 2. Fast-align provides efficient inference, likelihood evaluation, and parameter estimation, and is consistently faster than IBM Model 4. Its high-quality alignments have been shown to improve downstream tasks, making it well-suited for guiding sequence-to-sequence AEC models. 

\subsection{Feature Integration}

For phonetic feature integration, IPA features were embedded and combined with word embeddings through a multi-layer perceptron (MLP), allowing the Transformer to jointly leverage orthographic and phonetic representations. For alignment integration, we employed the alignment-assisted Neural Machine Translation (NMT) mechanism \cite{10319358}, where fast-align outputs were used as external constraints on the Transformer decoder’s attention distributions. Figure \ref{fig:framework} (b) shows the process of correcting ASR error using Phonetic (IPA) features and alignment information.

During training, these alignments guided and regularized attention weights, reducing spurious insertions or omissions and mitigating hallucinations. This supervision ensured that the model remained faithful to source–target correspondences while benefiting from the contextual modeling capacity of the Transformer.
\vspace{-1.7mm}
\subsection{Sequence-to-sequence Transformer}
We adopted the Transformer-based S2S architecture, in which, the encoder captures contextual representations of the input sequence through multi-head self-attention, while the decoder generates the corrected output sequence by attending to both the encoder states and previously generated tokens. This architecture is well-suited for AEC as it balances fluency and contextual modeling with parallelizable training. We trained a S2S Transformer model under different settings: with and without phonetic (IPA-based) features and, with and without alignment information. Training was conducted on both the original and the augmented datasets, and we compared model performance across these configurations.

\section{Experimental Setup}
For the ASR part, we used \texttt{transformer} library for Whisper finetuning and MMS pretraining inferencing. We used the \texttt{OpenNMT} toolkit \cite{klein-etal-2017-opennmt} for training the sequence-to-sequence Transformer models. CRF-suite was employed for grapheme-to-IPA (G2IPA) conversion, while fast-align was used for alignment extraction. 

For the Transformer architecture, both the encoder and decoder were configured with 4 layers, a hidden size of 512, and 512-dimensional word embeddings. The feed-forward network size was set to 2048 with 8 attention heads. Position encoding was enabled, and regularization included dropout (0.3) and attention dropout (0.3). Label smoothing was set to 0.1.

Training was performed with the Adam optimizer and the Noam learning rate schedule, with a learning rate of 0.1, 5{,}000 warm-up steps, and gradient accumulation of 4 steps. The maximum source and target sequence lengths were both limited to 200 tokens. The batch size was 64 (token-based normalization), and early stopping was applied with a patience of 4 validation checks to avoid overfitting. Models were trained for up to 200{,}000 steps, and the best 10 retained.

To assess the performance of our Transformer S2S based AEC models, we employed two widely used metrics: Word Error Rate (WER) and Character F-score (chrF{++}) \cite{popovic-2015-chrf}. WER captures the proportion of word-level substitutions, deletions, and insertions between the system output and the reference, making it a standard measure for correction accuracy at the lexical level. chrF{++} evaluates similarity at the character n-gram level, which is particularly useful for morphologically rich languages and for capturing finer-grained orthographic variations. Together, these metrics provide a balanced evaluation of both word-level accuracy and subword/character-level fidelity.


\begin{figure*}[ht!]
    \centering
    \caption{Average WER and chrF++ Score Across Different AEC Approaches Before and After Data Augmentation.}
    \includegraphics[width=\linewidth]{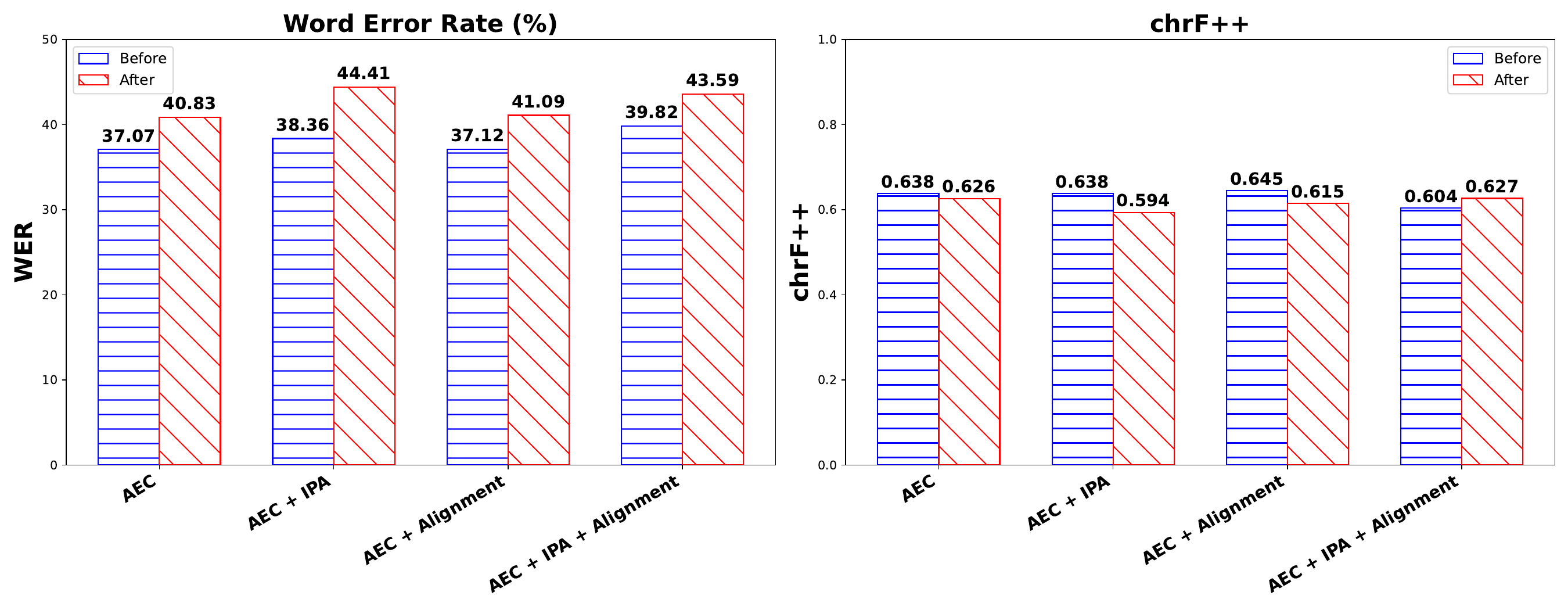}
     
    \label{fig:aec_wer_chrf}
\end{figure*}

\begin{table*}[ht!] 
\centering 
\renewcommand{\arraystretch}{1.2} 
\setlength{\tabcolsep}{6pt} 
\caption{Comparison of ASR Error Correction models trained with Different Features on both Original and Augmented Data Across Different ASR systems (Best results per ASR model in \textbf{Bold}).} 
\begin{tabular}{|c|l|c|c||c|l|c|c|} 
\hline \multicolumn{4}{|c||}{\textbf{Original Data}} & \multicolumn{4}{c|}{\textbf{+ Augmentation Data}} \\ 
\hline 
\textbf{ASR Model} & \textbf{Feature} & \textbf{WER} & \textbf{chrF{++}} & \textbf{ASR Model} 
& \textbf{Feature} & \textbf{WER} & \textbf{chrF{++}} \\ 
\hline 

\multirow{5}{*}{MMS} & No AEC (Baseline) & 42.27 & 0.6126 & \multirow{5}{*}{MMS} & No AEC (Baseline) & 42.27 & 0.6126 \\ & + AEC & 30.70 & 0.6461 & & + AEC & \textbf{33.76} & \textbf{0.6670} \\ & + AEC + IPA & 31.40 & 0.6923 & & + AEC + IPA & 38.85 & 0.6331 \\ & + AEC + Align & \textbf{30.21} & \textbf{0.6940} & & + AEC + Align & 34.07 & 0.6646 \\ & + AEC + IPA + Align & 35.03 & 0.6596 & & + AEC + IPA + Align & 36.57 & 0.6605 \\ 
\hline 
\multirow{5}{*}{Whisper Tiny} & No AEC (Baseline) & 55.07 & 0.5483 & \multirow{5}{*}{Whisper Tiny} & No AEC (Baseline) & 55.07 & 0.5483 \\ & + AEC & \textbf{43.79} & \textbf{0.5871} & & + AEC & \textbf{48.28} & \textbf{0.6245} \\ & + AEC + IPA & 45.21 & 0.5797 & & + AEC + IPA & 52.00 & 0.5361 \\ & + AEC + Align & 44.17 & 0.5868 & & + AEC + Align & 48.43 & 0.5546 \\ & + AEC + IPA + Align & 45.48 & 0.5463 & & + AEC + IPA + Align & 51.37 & 0.5816 \\ 
\hline 
\multirow{5}{*}{Whisper Small} & No AEC (Baseline) & 37.25 & 0.6534 & \multirow{5}{*}{Whisper Small} & No AEC (Baseline) & 37.25 & 0.6534 \\ & + AEC & \textbf{32.57} & 0.6745 & & + AEC & \textbf{36.22} & 0.6302 \\ & + AEC + IPA & 33.58 & 0.6709 & & + AEC + IPA & 38.57 & 0.6329 \\ & + AEC + Align & 32.71 & \textbf{0.6763} & & + AEC + Align & 36.27 & 0.6484 \\ & + AEC + IPA + Align & 33.71 & 0.6338 & & + AEC + IPA + Align & 38.23 & \textbf{0.6727} \\ \hline \multirow{5}{*}{Whisper Medium} & No AEC (Baseline) & 72.18 & 0.5425 & \multirow{5}{*}{Whisper Medium} & No AEC (Baseline) & 72.18 & 0.5425 \\ & + AEC & \textbf{41.88} & 0.6148 & & + AEC & \textbf{46.01} & 0.5884 \\ & + AEC + IPA & 43.50 & 0.6030 & & + AEC + IPA & 49.41 & 0.5629 \\ & + AEC + Align & 42.05 & \textbf{0.6169} & & + AEC + Align & 46.51 & 0.5844 \\ & + AEC + IPA + Align & 43.21 & 0.5616 & & + AEC + IPA + Align & 49.86 & \textbf{0.6041} \\ \hline \multirow{5}{*}{Whisper Large} & No AEC (Baseline) & 51.04 & 0.5752 & \multirow{5}{*}{Whisper Large} & No AEC (Baseline) & 51.04 & 0.5752 \\ & + AEC & \textbf{36.38} & \textbf{0.6666} & & + AEC & 39.87 & \textbf{0.6225} \\ & + AEC + IPA & 38.08 & 0.6425 & & + AEC + IPA & 43.21 & 0.6028 \\ & + AEC + Align & 36.47 & 0.6527 & & + AEC + Align & \textbf{40.17} & 0.6211 \\ & + AEC + IPA + Align & 41.66 & 0.6205 & & + AEC + IPA + Align & 41.94 & 0.6141 \\ 
\hline \end{tabular} 
\label{tab:wer_chrf_results} 
 
\end{table*}

\section{Result and Discussion}

\subsection{Across different AEC approaches}


Across all configurations, every AEC-based approach consistently outperforms the baseline ASR output without AEC, confirming the effectiveness of error correction in improving recognition accuracy. It reduces average WER of ASR models (ASR: 51.56 → AEC: 37.07) and improves average chrF++ scores (ASR: 0.5864 → AEC: 0.638). The results show that WER generally increased after augmentation across all feature setups (e.g., AEC: 37.07 → 40.83; AEC + Alignment: 37.12 → 41.09), reflecting the added difficulty of word-level prediction under augmented variability. Crucially, however, all AEC-based approaches (with or without augmentation) still outperform the baseline ASR output without AEC, confirming the robustness of error correction despite the distribution mismatch introduced by augmentation. Interestingly, chrF++ demonstrates a mixed effect. For some setups, such as AEC + Alignment + IPA, chrF++ increased after augmentation (0.6044 → 0.6266), suggesting that the augmented data helped the models produce outputs with better character-level overlap with references. However, other setups such as AEC + IPA experienced a drop (0.6377 → 0.5936). Overall, the averaged chrF++ indicates that augmentation can enhance surface-level similarity in certain configurations, particularly when alignment features are included. These findings suggest that while data augmentation may slightly degrade word-level accuracy, it can improve subword or character-level similarity in specific feature configurations, especially those leveraging alignment-based representations.

\subsection{Without Augmentation}
We first evaluated the ASR error correction models trained on the original dataset solely, without any augmented data. Across all ASR models, applying AEC yielded consistent improvements over the baseline (no AEC), confirming the robustness of error correction. Table \ref{tab:wer_chrf_results} reports the WER and chrF++ scores for each feature configuration across the different ASR models. The results show that AEC consistently improves performance over the ASR models without AEC. For example, MMS sees a reduction in WER from 42.27 to 30.70 and an increase in chrF++ from 0.6126 to 0.6461 when AEC is applied. Incorporating alignment-based features further improves performance, yielding the best overall results for MMS with a WER of 30.21 and chrF++ of 0.6940. IPA features have a mixed effect: they improve chrF++ for some models (e.g., MMS) but can slightly increase WER for others. Among the Whisper models, the Small and Large variants achieve the lowest WER and highest chrF++ with AEC and alignment, demonstrating that error correction and alignment features are particularly effective for stronger ASR backbones. These findings indicate that training on the original clean data is sufficient for large models to achieve high accuracy, and careful feature selection (AEC and alignment) can maximize both word-level and character-level performance.
 
\subsection{Effect of Data Augmentation}
 
We investigated the impact of augmenting the training data on ASR error correction across multiple ASR models. Table \ref{tab:wer_chrf_results} shows the WER and chrF++ scores for models trained on original data versus those trained with additional augmented data, evaluated on a fixed clean test set. Overall, the inclusion of augmented data consistently increased WER for all ASR models, indicating that the augmentation introduced variability that shifted the training distribution away from the clean test distribution. This distribution mismatch caused the models to perform worse at exact word-level recognition. In terms of chrF++, the effect of augmentation was more nuanced. For the smallest model, Whisper Tiny, chrF++ improved (0.5871 → 0.6245), suggesting that the augmented data helped the model generate outputs with higher character-level similarity to the references, even if full words were not always correct. For larger models (Whisper Small–Large, MMS), augmentation generally reduced chrF++, implying that these models, already strong on the original data, were negatively affected by the added variability. These results highlight that while augmentation can help under-parameterized models by providing additional variability, it may hinder performance for stronger models when evaluated on fixed clean data. Therefore, careful consideration is required when designing augmentation strategies, especially for high-performing ASR systems.

\begin{table*}[h!]
\centering
\scriptsize
\renewcommand{\arraystretch}{1.2}
\caption{Error Comparison of Different ASR Output (Errors are colored in \textcolor{red}{Red}).}
\begin{tabular}{|l|p{0.8\linewidth}|}
\hline
\textbf{ASR Models} & \textbf{Sentence} \\
\hline
Groundtruth & {\padauktext စိတ်|sei' ဝင်|win| စား|za: ဖွယ်|bwe ကောင်း|gaun: သော|tho: ရွာ|jwa သို့|dhou. အေး|ei: အေး|ei: လူ|lu လူ|lu ဖြင့်|hpjin. နာ|na ရီ|ji ဝက်|we' ခန့်|khan. လမ်း|lan: လျှောက်|shau' သွား|dhwa: ရ|ra$-$ သည်|dhe} \\
\hline
MMS & {\padauktext \textcolor{red}{အ|a- စတ်|ho} ဝင်|win စာ|za \textcolor{red}{ဘဲ|be: ကောင်|gaun တော်|do} ရွာ|jwa သို့|dhou. \textcolor{red}{a|x a|x} လူ|lu ဖြ|bu. င့်|bu. \textcolor{red}{နိာင်|bu.} ဝက်|we' ခန့်|khan. လမ်း|lan: လျှောက်|shau' သွား|dhwa: ရ|ra- သည်|dhe} \\
\hline
Whisper Tiny & {\padauktext \textcolor{red}{ဆိုက်|hsai'} ဝင်|win \textcolor{red}{ကြ|kya- ပဲ|pe:} ကောင်း|kaun: \textcolor{red}{လော်|lo ရာ|ja} သို့|dhou. \textcolor{red}{Aိတ်|x အေး|ei: ဘီ|bi} လူ|lu ဖြင့်|hpjin. \textcolor{red}{မ|ma- ဟုတ်|hou' ပွဲ|pwe: ခံ|gan နဲ|ne: ရှောက်|shau' ဘွား|hpwa: ရာ|ja} သည်|dhi
}\\
\hline
Whisper Small & {\padauktext စိတ်|sei' ဝင်|win စာ|sa- \textcolor{red}{ဘက်|be'} ကောင်း|kaun: \textcolor{red}{လောင်|laun} ရွာ|jwa သို့|dhou. အေး|ei: အေး|ei: လူ|lu ဖြင့်|hpjin. နာ|na ရီ|ji ဝက်|we' ခန့်|khan. လမ်း|lan: လျှောက်|shau' သွား|dhwa: ရ|ra- သည်|dhe}
\\
\hline
Whisper Medium & {\padauktext စိတ်|sei' ဝင်|win စား|za: \textcolor{red}{ဝဲ|we:} ကောင်း|gaun: \textcolor{red}{လော်|lo ရာ|ja သူး|htei အင်|in အီ|i ယူ|ju} ဖြင့်|hpjin. \textcolor{red}{နိုင်|nain} ဝက်|we' \textcolor{red}{ခံ|gan နှင်း|nhin: ရှောက်|shau'} သွား|thwa: ရ|ja. \textcolor{red}{လေ|lei}}
 \\
\hline
Whisper Large & {\padauktext စိတ်|sei' ဝင်|win စာ|sa \textcolor{red}{ပဲ|be:} ကောင်း|gaun: \textcolor{red}{လော်|lo ရ|ra' သူ|dhu ကျင်|gyin အေ|ei ဠူ|hta} ဖြင့်|hpjin. \textcolor{red}{နိုင်|nain} ဝက်|we' \textcolor{red}{ခံ|khan နှံ့|nhan.} လျှောက်|shau' သွား|thwa: ရ|ja. \textcolor{red}{လေ|lei}}\\
\hline
\end{tabular}
 
\label{tab:asr_error}
\end{table*}

\begin{table*}[h!]
\centering
\scriptsize
\renewcommand{\arraystretch}{1.2}
\caption{Error Comparison of Different AEC approaches for Whisper Tiny Model (Errors are colored in \textcolor{red}{Red}).}
\begin{tabular}{|l|p{0.8\linewidth}|}
\hline
\textbf{AEC Approaches} & \textbf{Sentence} \\
\hline
Correct & {\padauktext ရေ|jei စီး|zi: ကြောင်း|gyaun: မှ|mha. ထွက်|htwe' လာ|la သည်|thi နှင့်|ne. ရေ|jei ကူး|gu: ခြင်း|gyin: သည်|dhe ပုံ|boun မှန်|mhan အား|a: ဖြင့်|hpjin. မ|ma- ခက်|khe' ခဲ|ge: တော့|do. ပါ|ba} \\
\hline
Whisper Tiny & {\padauktext \textcolor{red}{ယ|ja-} စီး|si: \textcolor{red}{ချောင်း|chaun:} မှ|mha. ထွက်|htwe' လာ|la သည်|dhe  နှင့်|ne. ရေ|jei \textcolor{red}{ကို|kou ချင်|gyo သည်|thi} ပုံ|boun မှန်|mhan အား|a: \textcolor{red}{ပြင်|bjin} မ|ma- \textcolor{red}{ချက်|che' ခဲ့|shoun:} တော့|to. ပါ|ba} \\
\hline
+ AEC & {\padauktext ရေ|jei စီး|zi: ကြောင်း|gyaun: \textcolor{red}{မှာ|mha} ထွက်|dwe' လာ|la သည်|dhe  နှင့်|ne. ရေ|jei ကူး|gu: ခြင်း|gyin: သည်|dhe ပုံ|boun မှန်|mhan အား|a: ဖြင့်|hpjin. မ|ma- ခက်|khe' ခဲ|ge: \textcolor{red}{သော|tho:} ပါ|ba} \\
\hline
+ AEC + IPA & {\padauktext ရေ|jei စီး|zi: ကြောင်း|gyaun: မှ|mha. ထွက်|htwe' လာ|la သည်|thi နှင့်|ne. ရေ|jei ကူး|gu: ခြင်း|gyin: သည်|dhe ပုံ|boun မှန်|mhan အား|a: ဖြင့်|hpjin. မ|ma- ခက်|khe' ခဲ|ge: တော့|do. ပါ|ba} \\
\hline
+ AEC + Align & {\padauktext ရေ|jei စီး|zi: ကြောင်း|gyaun: \textcolor{red}{မှာ|mha} ထွက်|dwe' လာ|la \textcolor{red}{ပြီ|pji} နှင့်|ne. ရေ|jei ကူး|gu: ခြင်း|gyin: သည်|dhe ပုံ|boun မှန်|mhan အား|a: ဖြင့်|hpjin. မ|ma- ခက်|khe' ခဲ|ge: \textcolor{red}{သော|tho:} ပါ|ba}\\
\hline
+ AEC + IPA + Align & {\padauktext ရေ|jei စီး|zi: ကြောင်း|gyaun: မှ|mha. ထွက်|htwe' လာ|la သည်|thi နှင့်|ne. ရေ|jei ကူး|gu: ခြင်း|gyin: သည်|dhe ပုံ|boun မှန်|mhan အား|a: ဖြင့်|hpjin. မ|ma- ခက်|khe' ခဲ|ge: တော့|do. ပါ|ba} \\
\hline
\end{tabular}
 
\label{tab:post_asr_error}
\end{table*}

\section{Error Analysis}

%
\noindent
\textbf{ASR Error Analysis} Table~\ref{tab:asr_error} presents the recognition outputs of different ASR models compared against the ground truth. The analysis shows that all models capture the general sentence structure, but frequent misrecognitions appear at the word and syllable level. For example, Whisper Medium and Large often substitute semantically related but phonetically mismatched words. This indicates that larger models, despite their improved fluency, tend to hallucinate words when uncertain. Smaller models such as Whisper Tiny exhibit more severe lexical errors with unrelated segments. MMS shows inconsistent syllable alignment, producing non-standard transcriptions such as ``{\padauktext အ|a- စတ်|ho}" or repeated tokens (``a|x a|x"), reflecting unstable decoding on Myanmar text, which could hurt G2IPA conversion performance as well.

%
\noindent
\textbf{AEC Error Analysis} Table~\ref{tab:post_asr_error} compares different AEC approaches applied to Whisper Tiny output. 
The baseline Whisper Tiny transcription contains significant errors, e.g., ``{\padauktext ယ|ja-}" for ``{\padauktext ရေ|jei}" and ``{\padauktext ချောင်း|chaun:}" for ``{\padauktext ကြောင်း|gyaun:.}'' 
These lexical substitutions break semantic consistency. 
Applying AEC substantially reduces such errors, recovering correct forms of keywords and improving alignment with the ground truth. The plain AEC approach fixes major errors but sometimes introduces new ones, such as replacing ``{\padauktext မှ|mha.}" with ``{\padauktext မှာ|mha}" or ending with ``{\padauktext တော့|do.}" instead of ``{\padauktext သော|tho:}". The addition of IPA information (+AEC +IPA) enhances phonetic matching, producing almost identical output to the ground truth. Alignment constraints (+AEC +Align) further stabilize word order but may still insert wrong tokens like ``{\padauktext ပြီ|pji.}". 

\section{Conclusion and Future Work}
 
This work investigated ASR error correction (AEC) for low-resource Burmese across different feature configurations and training setups. Our findings show that all AEC-based approaches, regardless of augmentation, consistently outperform the baseline ASR outputs without AEC, highlighting the effectiveness of error correction in improving recognition quality. Alignment features provided the strongest gains, especially when combined with AEC, while IPA features yielded mixed effects depending on the backbone ASR model. Data augmentation was found to introduce variability that increased WER on clean test data, although it sometimes improved character-level similarity (chrF++) for smaller models. This suggests that augmentation can be beneficial in under-parameterized scenarios but may lead to distribution mismatch for stronger ASR backbones. Overall, these results underscore the robustness of AEC and the importance of careful feature design. Future work will explore augmentation strategies better matched to test conditions, as well as extending AEC to multimodal scenarios \cite{mu2024mmgermultimodalmultigranularitygenerative} and integrating large language models \cite{sachdev2024evolutionarypromptdesignllmbased} for enhanced correction capabilities.

For future work, we plan to release the fine-tuned Whisper models and AEC models, including \texttt{OpenNMT} configuration files, along with the parallel corpus of ASR outputs, ground-truth and their corresponding AEC-corrected transcripts, to support further study in low-resource Burmese ASR error correction.

\addtolength{\textheight}{0.2in}
\newpage
\bibliographystyle{IEEEtran}
\bibliography{references}

\end{document}